\documentclass[runningheads]{paper1034-llncs}
\usepackage[T1]{fontenc}
\usepackage{graphicx}
\usepackage{graphicx}
\usepackage{multirow} 
\usepackage{float}
\usepackage[para]{threeparttable}
\usepackage{adjustbox}
\usepackage{amsmath}
\usepackage{bm}

\usepackage{arydshln}
\usepackage{amssymb}
\usepackage{xr}
\usepackage{amsfonts}
\usepackage[misc]{ifsym}

\usepackage{color}
\newcommand\blfootnote[1]{%
  \begingroup
  \renewcommand\thefootnote{}\footnote{#1}%
  \addtocounter{footnote}{-1}%
  \endgroup
}
\begin{document}
\title{Lesion Guided Explainable Few Weak-shot Medical Report Generation}
%
%
\author{Jinghan Sun\inst{1,2} \and
Dong Wei\inst{2} \and
Liansheng Wang\inst{1}\textsuperscript{(\Letter)} \and
Yefeng Zheng\inst{2}}
%
%
\authorrunning{J. Sun et al.}
%
%
\institute{Xiamen University, Xiamen, China\\
\email{jhsun@stu.xmu.edu.cn}, \email{lswang@xmu.edu.cn} \and
Tencent Healthcare (Shenzhen) Co., LTD, Tencent Jarvis Lab, Shenzhen, China\\
\email{\{donwei,yefengzheng\}@tencent.com}}
\maketitle              
\begin{abstract}
Medical images are widely used in clinical practice for diagnosis\blfootnote{\textsuperscript{*} J. Sun and D. Wei---Contributed equally; J. Sun contributed to this work during an internship at Tencent.}.
Automatically generating interpretable medical reports can reduce radiologists' burden and facilitate timely care. However, most existing approaches to automatic report generation require sufficient labeled data for training. 
In addition, the learned model can only generate reports for the training classes, lacking the ability to adapt to previously unseen novel diseases. 
To this end, we propose a lesion guided explainable few weak-shot medical report generation framework that learns correlation between seen and novel classes through visual and semantic feature alignment, aiming to generate medical reports for diseases not observed in training. 
It integrates a lesion-centric feature extractor and a Transformer-based report generation module. 
Concretely, the lesion-centric feature extractor detects the abnormal regions and learns correlations between seen and novel classes with multi-view (visual and lexical) embeddings. 
Then, features of the detected regions and corresponding embeddings are concatenated as multi-view input to the report generation module for explainable report generation, including text descriptions and corresponding abnormal regions detected in the images.
We conduct experiments on FFA-IR, a dataset providing explainable annotations, showing that our framework outperforms others on report generation for novel diseases.

\end{abstract}

\section{Introduction}

\begin{figure}[t]
\centering
\includegraphics[width=.6\textwidth]{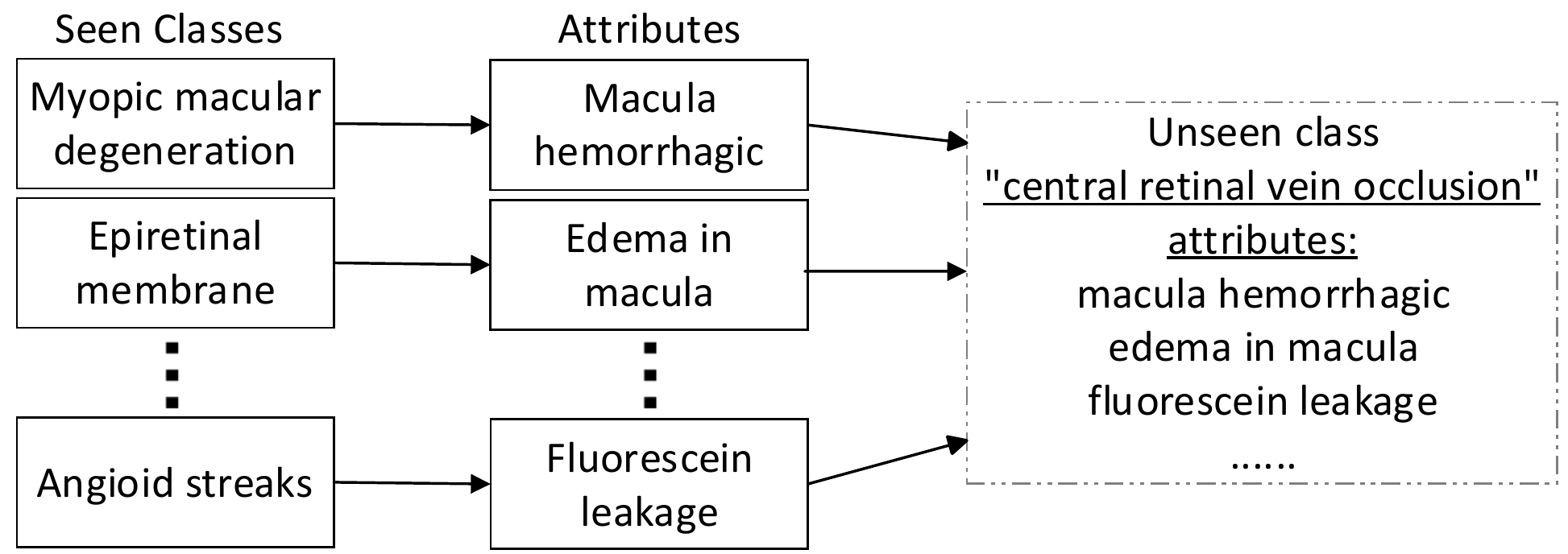}
\caption{Illustration of our motivation:
the seen classes are used to learn a mapping between visual and lexical features in the training stage, which is then transferred to generate reports for previously unseen novel classes.}
\label{fig:example}
\end{figure}

Medical images are widely used in clinics for diagnosis, prognosis, and therapy planning. 
Radiologists usually write imaging reports based on their expertise and experience while observing abnormal areas in the images. 
However, due to the demand for expert knowledge and large number of images, this process is laborious, time-consuming, and error-prone.
To relieve this burden, a large body of literature \cite{chen2020generating,jing2019show,johnson2019mimic,liu2019clinically,xue2018multimodal} has investigated automatic radiology report generation.
However, there are still some practical limitations of these methods. 
A notable one is that the trained generator can only generate reports for the training classes and the performance will degrade on novel diseases.
Several few-shot learning (FSL) report generation methods \cite{jia2021radiology,jia2020few} have been proposed to tackle this problem, which require the ground truth reports of the novel diseases to fine-tune the network.
However, in clinical practice, ground truth reports for the novel diseases can be unavailable, too.

To enable the model to recognize previously unseen classes, zero-shot learning (ZSL) \cite{larochelle2008zero} is proposed for natural image tasks.
In ZSL \cite{han2021contrastive,jiang2019transferable,xian2017zero}, a mapping function is learned to project visual input to a semantic space, where the relationship between seen and novel classes is extracted from an auxiliary source, e.g., word embeddings.
Being inspired, 
we propose to build a connection between seen training diseases and novel testing diseases through learning common attributes. 
Taking Fig. \ref{fig:example} for example, although in the training stage the model does not see a report of the novel class---central retinal vein occlusion, we can make it learn a mapping from visual features to semantic features, with which the correlation between seen and novel diseases can be established.
To ensure a well alignment of both seen and novel classes' semantic features, we propose a multi-view embedding ensemble strategy where lexical embeddings \cite{wang2019survey} of the diseases are employed for feature calibration.
Then, when a novel disease sample is presented, the model can make use of the calibrated correlation for report generation by assembling descriptions of the common attributes.
However, ZSL approaches often cannot produce optimal results, 
as the learned relationship between seen and novel classes may be weak and ambiguous.

Alternatively, some researchers propose to provide weak annotations for the novel classes, which are more accessible than full annotations and meanwhile can provide more definite information about the novel classes than ZSL.
This sort of task is known as weak-shot learning \cite{chen2021weak,zhou2021weak}.
However, for the task of quickly adapting to generate reports for novel diseases, collecting enough weakly annotated samples can be prohibitive, too.
Thus, it is desirable to allow \textit{few} weak-shot learning in this scenario.

In this work, we propose multi-view lesion guided few weak-shot learning for explainable medical report generation. 
Concretely, we first extract regional lesion features based on Faster R-CNN \cite{ren2015faster} to guide the network to capture the noteworthy features and output detection results to provide explainable prediction for reports. 
In addition, we adopt the weight imprinting strategy \cite{qi2018low} based on a few novel samples that are weakly annotated with labeled bounding boxes to capture novel class features in the testing stage. Then, we project the visual features into semantic space and construct a soft target of all classes. 
The soft target indicates the correlation between the lexical embeddings of seen classes and novel classes, which force the network to learn about the novel classes although only seen classes are available for training. 
Finally, the visual features and lexical embeddings are concatenated as multi-view features and fed into the generation model to generate reports.
In summary, the contributions of this paper reside in:
(1) This is the first study that proposes few weak-shot medical report generation, which can generate explainable reports of novel classes without any training report.
(2) We extract lesion proposal features instead of global visual features to guide the model for more focused report generation.
(3) We propose a soft target of lexical embeddings, through which features of previously unseen classes can be effectively learned. 
Furthermore, visual features and lexical embeddings are concatenated as the input to provide multi-view knowledge for the generation model.
(4) We conduct experiments on the Fundus Fluorescein Angiography Images and Reports dataset (FFA-IR) \cite{li2021ffa} with six experimental settings. 
The results demonstrate the superiority of the proposed method.

\section{Methods}
\subsubsection{Problem Setting.}
We denote the set of disease classes as $C = C^\mathrm{seen} \cup C^\mathrm{nov}$, where $C^\mathrm{seen}$ and $C^\mathrm{nov}$ denote the sets of seen and novel diseases, respectively, satisfying $C^\mathrm{seen} \cap C^\mathrm{nov} = \emptyset$. 
Each case in the dataset contains images of a patient at different periods: $\{I_1,\ldots,I_N\}$, where $N\geq 1$ is the total number of images in one case. Notably, $N$ is not a parameter but a property of the data;
the proposed method can also generalize for non-periodical data, e.g., $N$ views in a chest X-ray.
For seen classes, explainable annotations including a set of labeled bounding boxes $B=\{(X, y)\}$ for the images and a report $R=\{r_1,\ldots,r_T\}, r_t\in \mathbb{V}$ are provided for each case, where $X$ specifies the location and size of each box, $y$ is the disease class label and $y \in \{1, \ldots, |C^\mathrm{seen}|\}$, $T$ is the length of reports, $\mathbb{V}$ is a vocabulary of all possible tokens. 
For novel diseases, we sample $K$ images for each class to compose the support set $S$, with all remaining images composing the query set $Q$.
Different from existing weak-shot methods that provide weak annotations for all samples of the novel classes, we only provide weak annotations (i.e., labeled bounding boxes $B$ in this work) for the few samples in the support set $S$.
It is worth noting that our architecture also has the potential to use other weak annotations, which we leave for future work.
Given $S$, the target is to generate explainable medical reports for the query set $Q$ by transferring the explainable knowledge of seen diseases.
To this end, we propose a novel few weak-shot task for medical report generation.

\begin{figure}[t]
\centering
\includegraphics[width=\textwidth]{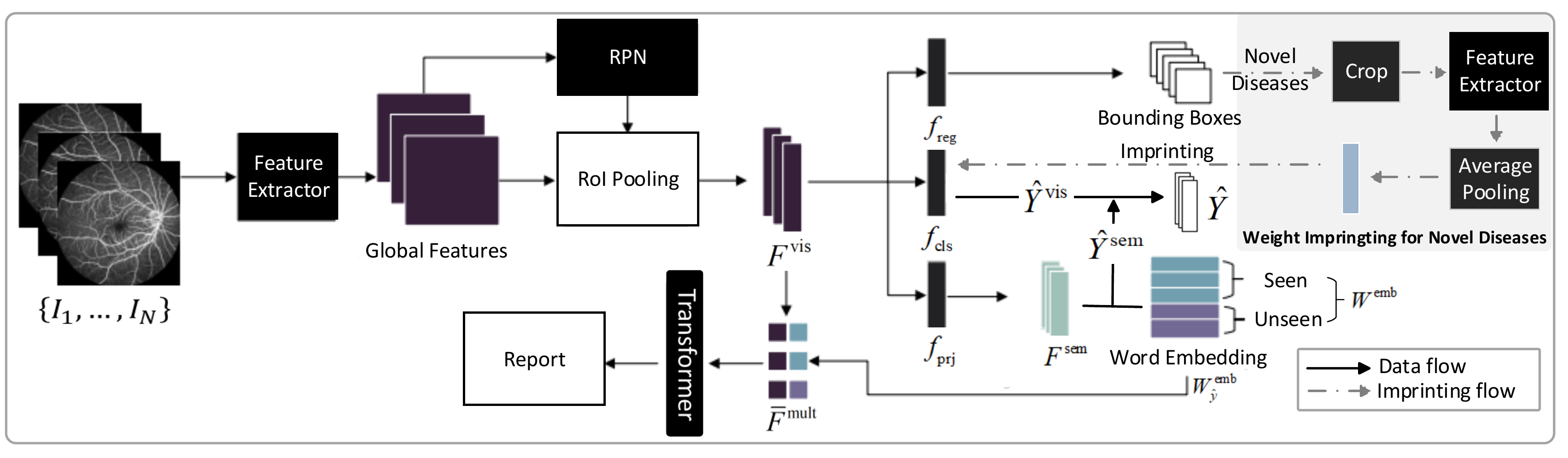}
\caption{Overview of the proposed approach.}
\label{fig:procedure}
\end{figure}

\subsubsection{Method Overview.}
Fig. \ref{fig:procedure} shows an overview of our method. 
Given the images of a patient, we first extract global features from each image $I$. 
Then the region proposal network (RPN) generates proposals of lesion regions and uses region of interest (ROI) pooling to extract regional visual features $F^\mathrm{vis}$. 
$F^\mathrm{vis}$ are sent into three branches: 1. box regression branch ($f_\mathrm{reg}$) to detect the location and size of lesion regions; 2. classification branch ($f_\mathrm{cls}$) to predict category of each region; 3. semantic prediction branch ($f_\mathrm{prj}$) to project visual features to semantic features ($F^\mathrm{sem}$) and learn the relationship between seen and novel diseases. Finally, we concatenate visual features and their corresponding word embeddings, and send the concatenated features $\bar{F}^\mathrm{mult}$ into Transformer \cite{vaswani2017attention} to generate the report. In the inference stage, we employ the weight imprinting technique (top right of Fig. \ref{fig:procedure}) to enable $f_\mathrm{cls}$ to recognize novel diseases given the few weak-shot samples.

\subsubsection{Lesion-Centric Feature Extractor.}
The key ingredient of report generation is the performance of extracted visual features. However, existing report generation frameworks \cite{chen2020generating,jing2019show} usually use global features for the generation model, which impedes the quality of generated reports. 
We propose to extract features from bounded lesion regions to provide the report generation model with more representative knowledge.
Our visual extractor is built on Faster R-CNN \cite{ren2015faster}, a popular object detection algorithm. 
In the first stage, Faster R-CNN extracts a global feature for an image. In the second stage, the region proposal network generates lesion region proposals. 
Then the network uses ROI pooling to extract fixed-size visual feature $F^\mathrm{vis}$ for each region proposal. These features are fed into two fully connection layers---a box regression layer $f_\mathrm{reg}$ and a box classification layer $f_\mathrm{cls}$ parameterized by $\theta^\mathrm{reg}$ and $\theta^\mathrm{cls}$, respectively.
For $f_\mathrm{cls}$, we adopt the weight imprinting mechanism \cite{qi2018low} to enable few weak-shot report generation (described later), and enforce $\left \| F^\mathrm{vis}\right\|_2=1$ via L2 normalization. 
The weight matrix of $\theta^\mathrm{cls}$ can be viewed as a learnable visual embedding dictionary of the diseases. 
Then, the detection loss is defined as:
\begin{equation}
\mathcal{L}_\mathrm{det}= 
\mathcal{L}_\mathrm{cls}\left((\hat{Y}^\mathrm{vis}+\hat{Y}^\mathrm{sem})/2, y\right) 
+ \mathcal{L}_\mathrm{reg}\left(f_\mathrm{reg}(F^\mathrm{vis}), X\right),
\end{equation}
where $\hat{Y}^\mathrm{vis}=[\hat{p}^\mathrm{vis}_1, \ldots, \hat{p}^\mathrm{vis}_{|C^\mathrm{*}|}]$ is the visual prediction by $f_\mathrm{cls}$ after softmax, $C^\mathrm{*}$ must be $C^\mathrm{seen}$ for training with seen diseases but can be any of $C^\mathrm{seen}$, $C^\mathrm{nov}$, or $C$ in different test settings, 
$\hat{Y}^\mathrm{sem}$ is the semantic prediction (described below), $y$ is the class label, and $X$ is the location and size of each box. The classification loss $\mathcal{L}_\mathrm{cls}$ is the cross-entropy loss. For regression, we adopt the smooth L1 loss: $\mathcal{L}_\mathrm{reg}=0.5 x^{2}$  if $|x|<1$ and $|x|-0.5$  otherwise. 

\subsubsection{Multi-view Embedding Ensemble.}
Report generation is an image-to-text task, so visual and semantic features both contain valuable information to improve report generation. 
The Faster R-CNN is trained on seen diseases but neglects the contents of novel diseases.
As the reports of novel diseases are not available, we can learn from the lexical embeddings of novel diseases to guide the visual feature learning.
To make the model transferable over novel diseases through learning relationship between seen and novel diseases, we first create a semantic space and project visual feature $F^\mathrm{vis}$ into this semantic space via linear projection: 
$F^\mathrm{sem}=f_\mathrm{prj} (F^\mathrm{vis})=\theta^\mathrm{prj} F^\mathrm{vis}$, where $f_\mathrm{prj}$ is a projection function parameterized by $\theta^\mathrm{prj}$. 
Then we obtain lexical embeddings $W^\mathrm{emb} \in \mathbb{R}^{|C|\times d}$ of both seen and novel diseases from a pretrained word embedding model such as BioBert \cite{lee2020biobert}, where $d$ is the dimension of each embedding. 
Instead of directly maximizing the similarity between visual features and their corresponding lexical embeddings, we build a soft label for exploiting the semantic relationship between seen and novel diseases. Specifically, we define the soft label for class $y$ as: 
\begin{equation}
L^\mathrm{s}_y=[l^\mathrm{s}_1, \ldots, l^\mathrm{s}_{|C|}]^T, \text{ where } l^\mathrm{s}_c=
\operatorname{Sim}(W_y^\mathrm{emb}, W_c^\mathrm{emb}),
\end{equation}
and $\operatorname{Sim}$ is the cosine similarity function to measure the similarity of class $y$ to class $c$ in the semantic space.
Thus, $L^\mathrm{s}_y$ collects class $y$’s similarity to all classes, both seen and novel, establishing correlations between them.
The loss function in the semantic space is $\mathcal{L}_\mathrm{kl}(\operatorname{Sim}(F^\mathrm{sem},W^\mathrm{emb}),L^\mathrm{s})$, where $\mathcal{L}_\mathrm{kl}$ is the Kullback-Leibler divergence loss. 
In addition, we define $\hat{Y}^\mathrm{sem}=\operatorname{softmax}(\operatorname{Sim}(F^\mathrm{sem},W^\mathrm{emb}))$ 
as the semantic prediction to assist the detection
(Eqn. (1)). 
Thus, we not only force the network to learn the relationship between seen and novel diseases through $\mathcal{L}_\mathrm{kl}$, but also boost object detection with the semantic information. 

\subsubsection{Lesion Guided Report Generation.}
Conventional report generation approaches usually treated the lesions and non-lesion regions equally. 
In contrast, we propose to merge lesion guided visual and semantic features to improve report generation. 
Concretely, given a set of lesion proposals $\hat{B}$ predicted in all images of a patient, and the corresponding visual features $\{F^\mathrm{vis}\}$ and labels $\{\hat{Y}\}$, 
where $\hat{Y}=(\hat{Y}^\mathrm{vis}+\hat{Y}^\mathrm{sem})/2=[\hat{p}_1,\ldots, \hat{p}_{|C^\mathrm{*}|}]$ and $\sum_{c=1}^{|C^\mathrm{*}|}\hat{p}_c=1$,
the corresponding set of class predictions can be obtained by $\{\hat{y}=\operatorname{argmax}_c\hat{p}_c\}$.
Then, we obtain the corresponding lexical embeddings $\{W^\mathrm{emb}_{\hat{y}}\}$.
Next, a multi-view feature is obtained by concatenating the visual feature and lexical embedding $F^\mathrm{mult}=[F^\mathrm{vis}, W^\mathrm{emb}_{\hat{y}}]$.
Finally, the input feature to the report generator is obtained as the average multi-view feature over all lesion proposals, weighted by each proposal's prediction probability:
$\bar{F}^\mathrm{mult}=\frac{1}{|\hat{B}|}{\sum}_{b=1}^{|\hat{B}|} \operatorname{max}(\hat{Y}_b)F_b^\mathrm{mult}.$

For report generation, we employ the commonly used Transformer \cite{amjoud2021automatic,li2021ffa,vaswani2017attention} architecture.
Denoting the report generator by $f_\mathrm{trans}$, then we have $\{\hat{r}_{1}, \ldots, \hat{r}_{T}\}=f_\mathrm{trans}\left(\bar{F}^\mathrm{mult}\right)$.
The loss function is defined as: 
\begin{equation}
\mathcal{L}_\mathrm{gen}=-\log p\left(\hat{r}_{T} \mid \hat{r}_1, \ldots, \hat{r}_{T-1}; \theta^\mathrm{trans} \right),
\end{equation}
where $\theta^\mathrm{trans}$ are the parameters of of the Transformer.
For more details of the Transformer, we refer readers to \cite{vaswani2017attention}. 

The overall optimization objective of our method is defined as:
\begin{equation}
\mathcal{L}=\mathcal{L}_\mathrm{det}/|\hat{B}| + \mathcal{L}_\mathrm{kl}/|\hat{B}| + \mathcal{L}_\mathrm{gen}.
\end{equation}

\subsubsection{Few Weak-Shot Report Generation for Novel Diseases.}
In the inference stage, to ensure that the model can accurately recognize previously unseen novel diseases with only few weakly labeled examples in the support set $S$, we employ the weight imprinting scheme \cite{qi2018low}. 
Concretely, given a few examples for a novel disease $c$, we first make use of the box annotations $B$ to crop the lesion regions for each class. 
Then, we obtain the visual features $\{F^\mathrm{vis}_b\}_{b=1}^{|B|}$ by feeding the cropped regions to the global visual extractor (without using the RPN or ROI pooling).
We compute new weights for the classification layer ($f_\mathrm{cls}$) by averaging the normalized visual features: $\theta^\mathrm{cls}_c=(\frac{1}{|B|} \sum F^\mathrm{vis}_b)/\big \| \frac{1}{|B|} \sum F^\mathrm{vis}_b \big\|_2$, where $\theta^\mathrm{cls}_c$ is the classification weights for the novel class $c$.
As discussed in \cite{qi2018low}, the embeddings of novel classes may not have a unimodal distribution and influence the detection accuracy. 
Therefore, we conduct fine-tuning to diminish the bias in the learned embedding space after the weight embedding.
To generate explainable reports for novel classes in the query set $Q$, we first predict the bounding boxes and classification results (both visual and semantic)
through forward propagation and then apply non-maximum suppression with a threshold of 0.5 to obtain the final detections. 
Then the visual features and corresponding lexical embeddings are concatenated and input to the Transformer to generate reports.

\section{Experiments}
\subsubsection{Dataset and Evaluation Metrics.}
FFA-IR \cite{li2021ffa} is a new benchmark based on fundus fluorescein angiography images and reports. FFA-IR includes annotations of 46 categories of lesions including 315 cases with 12,166 lesion regions. 
For each case, we use the English reports, lesion category, lesion regions and the images. 
We conduct experiments with two different splits of train (seen), validation and test (novel) classes of 32/5/9 and 34/5/7 classes, respectively, to verify our method's robustness to different task scenarios.
For the training set, we first select all images containing lesions of seen diseases.
Then, we exclude those images with any lesion of the novel diseases to make sure the novel classes are not observed during training stage.
For testing, $K$ is set to five (i.e., five examples for each novel disease class) for the support set $S$. 
We use three different settings following \cite{zhu2019zero}: Test-Seen, Test-Novel and Test-Mix.
Test-Seen data contain images of only seen diseases, 
Test-Novel 
contain images of only novel diseases, and Test-Mix contain images of both seen and novel classes---which is a more challenging task known as generalized weak shot.
We employ seven commonly used metrics to evaluate the quality of generated reports, including: BLEU (1- to 4-gram) \cite{papineni2002bleu}, CIDER \cite{vedantam2015cider}, METEOR \cite{banerjee2005meteor}, and ROUGE \cite{lin2004rouge}. 

\begin{table}[t]
\centering
\footnotesize
\caption{The performance of our framework and exiting SOTA methods. B*N denotes N-gram score of BLEU. Best results are shown in bold.}
\label{tab:sota}
\begin{adjustbox}{width=0.7\columnwidth}
\begin{tabular}{lllccccccc}
\hline
                        & \multicolumn{1}{c}{Split}                 & Model    & B1             & B2             & B3             & B4             & METEOR         & ROUGE          & CIDER          \\ \hline
\multirow{6}{*}{Test-Seen}   & \multirow{3}{*}{32/5/9}                     & Grounded \cite{zhou2020more} & 0.477          & 0.346          & 0.253          & 0.190          & \textbf{0.233} & \textbf{0.407} & 0.511          \\
                        &                                           & R2Gen \cite{chen2020generating}    & 0.472          & 0.347          & 0.247          & 0.184          & 0.215          & 0.402          & 0.460          \\
                        &                                           & Ours     & \textbf{0.491} & \textbf{0.357} & \textbf{0.263} & \textbf{0.200} & 0.223          & 0.398          & \textbf{0.519} \\ \cline{2-10} 
                        & \multirow{3}{*}{34/5/7}                     & Grounded \cite{zhou2020more} & 0.494          & 0.356          & \textbf{0.262} & 0.197          & 0.220          & \textbf{0.410} & 0.558          \\
                        &                                           & R2Gen \cite{chen2020generating}    & 0.459          & 0.311          & 0.222          & 0.171          & 0.222          & 0.383          & 0.541          \\
                        &                                           & Ours     & \textbf{0.511} & \textbf{0.357} & \textbf{0.262} & \textbf{0.199} & \textbf{0.223} & 0.401          & \textbf{0.571} \\ \hline
\multirow{6}{*}{Test-Novel} & \multirow{3}{*}{32/5/9}                     & Grounded \cite{zhou2020more} & 0.466          & 0.341          & 0.239          & 0.183          & \textbf{0.220} & 0.375          & 0.472          \\
                        &                                           & R2Gen \cite{chen2020generating}    & 0.434          & 0.324          & 0.232          & 0.175          & 0.208          & 0.374          & 0.331          \\
                        &                                           & Ours     & \textbf{0.481} & \textbf{0.349} & \textbf{0.256} & \textbf{0.193} & 0.217          & \textbf{0.409} & \textbf{0.496} \\ \cline{2-10} 
                        & \multirow{3}{*}{34/5/7}                     & Grounded \cite{zhou2020more} & 0.435          & 0.348          & 0.257          & 0.212          & \textbf{0.230} & \textbf{0.417} & 0.551          \\
                        &                                           & R2Gen \cite{chen2020generating}    & 0.402          & 0.320          & 0.211          & 0.209          & 0.209          & 0.403          & 0.492          \\
                        &                                           & Ours     & \textbf{0.510} & \textbf{0.371} & \textbf{0.276} & \textbf{0.217} & 0.228          & 0.404          & \textbf{0.564} \\ \hline
\multirow{6}{*}{Test-Mix}    & \multicolumn{1}{c}{\multirow{3}{*}{32/5/9}} & Grounded \cite{zhou2020more} & 0.453          & 0.343          & \textbf{0.267} & \textbf{0.207} & 0.174          & 0.392          & 0.507          \\
                        & \multicolumn{1}{c}{}                      & R2Gen \cite{chen2020generating}    & 0.464          & 0.321          & 0.239          & 0.183          & 0.175          & 0.375          & 0.412          \\
                        & \multicolumn{1}{c}{}                      & Ours     & \textbf{0.490} & \textbf{0.356} & 0.262          & 0.199          & \textbf{0.222} & \textbf{0.398} & \textbf{0.511} \\ \cline{2-10} 
                        & \multicolumn{1}{c}{\multirow{3}{*}{34/5/7}} & Grounded \cite{zhou2020more} & 0.426          & 0.297          & 0.218          & 0.170          & 0.210          & \textbf{0.386} & 0.535          \\
                        & \multicolumn{1}{c}{}                      & R2Gen \cite{chen2020generating}    & 0.456          & 0.309          & \textbf{0.224} & 0.163          & 0.208          & 0.384          & 0.445          \\
                        & \multicolumn{1}{c}{}                      & Ours     & \textbf{0.459} & \textbf{0.311} & 0.222          & \textbf{0.171} & \textbf{0.222} & 0.383          & \textbf{0.541} \\ \hline
\end{tabular}
\end{adjustbox}
\end{table}

\subsubsection{Implementation.}
PyTorch \cite{paszke2019pytorch} (1.4.0) is used for experiments. We use ResNet-101 \cite{he2016deep} pretrained on ImageNet \cite{deng2009imagenet} as the visual extractor of Faster R-CNN. We train the network for 100 epochs with a mini-batch of four cases on four Tesla V100 GPUs. 
The SGD optimizer is used with a momentum 0.9 and a decay of 0.0005. 
We set initial learning rate to $10^{-4}$ and decay at 60 and 80 epochs by multiplying by 0.1, respectively. 
All images are resized to 224$\times$224 pixels. 
For the rest, we follow the optimal training strategies suggested in \cite{li2021ffa}.
The source code is available at: https://github.com/jinghanSunn/Few-weak-shot-RG.

\begin{figure}[t]
\centering
\includegraphics[width=0.9\textwidth]{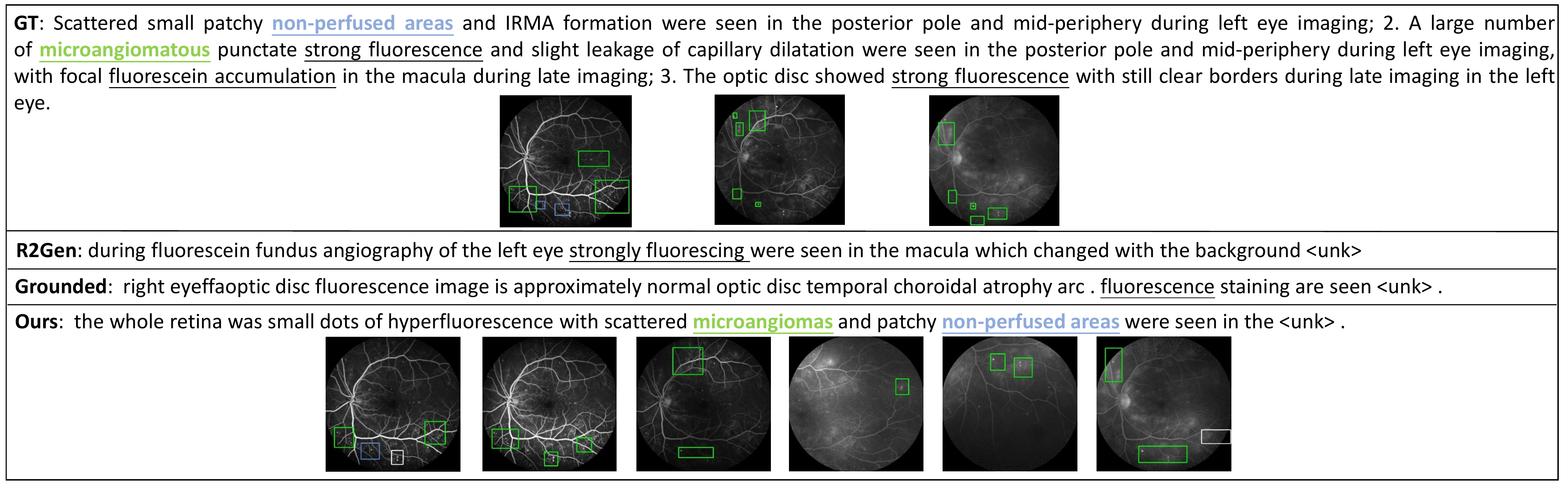}
\caption{The visualization of ground truth report (GT), and reports generated by R2Gen \cite{chen2020generating}, Grounded \cite{zhou2020more} and ours.
The green and blue boxes are lesion regions of two novel classes. 
The white boxes in our results are predicted abnormalities that do not belong to these two classes.
}
\label{fig:visualization}
\end{figure}

\subsubsection{Comparison to State-of-the-Art (SOTA) Methods.}
We compare our met-hod with Grounded \cite{zhou2020more} and R2Gen \cite{chen2020generating}, for their excellent performance in medical report generation.
In table \ref{tab:sota}, we can see that our method outperforms existing SOTA methods in most metrics (5 to 6 of 7 metrics) on both split configurations and three test settings. 
The results demonstrate the effectiveness of our method for the few weak-shot report generation task.
The performance of the 32/5/9 split is slightly worse than that of 34/5/7, because the 32/5/9 split task is more difficult as the number of training data and diversity are reduced while more novel classes need to be discriminated.
Fig. \ref{fig:visualization} shows example reports generated by our method, R2Gen \cite{chen2020generating} and Grounded \cite{zhou2020more}, from which we make the following observations. First, in the ground truth report, only three images have labeled bounding boxes which may cause a great loss of critical diagnostic information. 
Second, R2Gen \cite{chen2020generating} and Grounded \cite{zhou2020more} cannot generate satisfactory description of novel classes. 
They only perceives non-critical knowledge, such as ``strongly fluorescing''. 
Last, our method generates a report including descriptions of the novel diseases, with corresponding lesion regions detected in images of various periods for explainable reporting.

To further validate the efficacy of our method with varying $K$ values, we conduct experiments with the Test-Novel setting on the test set of 34/5/9 split (supplement Table 1).
Results show that larger $K$ values generally yield better results, as expected, and that our method overall outperforms Grounded and R2Gen with different $K$ values.

\subsubsection{Ablation Study.}
We conduct ablation studies to validate the effects of our proposed components on the validation set of the 34/5/7 split. 
From the results in Table \ref{tab:ablation}, we make the following observations. 
First, we replace the Faster R-CNN based visual extractor with ResNet-101 \cite{he2016deep} (Ablation-1). 
Results show the Faster R-CNN based visual extractor (Ablation-2) achieves better performance to vanilla CNN (Ablation-1) as Faster R-CNN can capture lesion-centric features. 
Second, equipping with imprinted weights (Ablation-3) assists the model to capture novel diseases' features, improving the generation performance. 
Third, combining semantic embeddings with visual features (Ablation-4) outperforms single-view method in four of the seven metrics. 
However, it is worse than our complete model as Ablation-4 lacks the ability to learn the relationship via soft labels, with which the correlation between seen and novel diseases can be established. 
With all these components, our complete model yields the best overall performance, achieving the best results for six metrics and the second best for the remaining one metric.

\begin{table}[t]
\caption{Ablation study on validation set with the 34/5/7 split. B*N denotes N-gram score of BLEU \cite{papineni2002bleu}. Best results are shown in bold.}
\label{tab:ablation}
\centering
\begin{adjustbox}{width=0.8\columnwidth}
\begin{tabular}{lccccccccccc}
\hline
           & Faster R-CNN  & Imprinted & Multi-view & Soft label  & B1             & B2             & B3             & B4             & METEOR         & ROUGE          & CIDER          \\ \hline
Ablation-1 &           &           &            &           & 0.459          & 0.306          & 0.261          & 0.231          & 0.211          & 0.410          & 0.507          \\
Ablation-2 & \checkmark &           &            &           & 0.488          & 0.323          & 0.262          & 0.232          & 0.197          & 0.387          & 0.532          \\
Ablation-3 & \checkmark & \checkmark &            &           & 0.508          & 0.339          & 0.274          & 0.242          & 0.250          & \textbf{0.412} & 0.560          \\
Ablation-4 & \checkmark & \checkmark & \checkmark  &           & 0.513          & 0.377          & 0.279          & 0.223          & 0.229          & 0.389          & 0.565          \\
Ours       & \checkmark & \checkmark & \checkmark  & \checkmark & \textbf{0.524} & \textbf{0.389} & \textbf{0.280} & \textbf{0.251} & \textbf{0.254} & 0.411          & \textbf{0.568} \\  \hline
\end{tabular}
\end{adjustbox}
\end{table}

\section{Conclusion}
In this work, we introduced a novel few weak-shot report generation task, where no ground truth report but only a few box-annotated image samples of novel diseases were available. 
To tackle this new task, we proposed a multi-view lesion guided report generation framework. 
We first detected lesion regions to extract lesion-centric visual features and output abnormal regions.
Then we projected visual features into semantic space where a word embedding model was utilized to learn the relationship between seen and novel classes. 
Finally, we combined the multi-view features for report generation. 
The reports were accompanied by detected abnormal regions in the input images to provide enhanced explainability.
Our method generally outperformed other state-of-art approaches to medical report generation in most metrics (5 to 6 of 7 metrics) and six experimental settings.
In the future, we plan to employ more advanced few-shot detection methods to improve the absolute scores.

%
%
%
\subsubsection{Acknowledgement.} This work was supported by the Scientific and Technical Innovation 2030 - ``New Generation Artificial Intelligence'' Project (No. 2020AAA0104100) and the National Key Research and Development Program of
China (2019YFE0113900).

%

\end{document}